\documentclass[10pt,twocolumn,letterpaper]{article}

\usepackage{cvpr}              
\usepackage{colortbl}  

\usepackage[dvipsnames]{xcolor}

\definecolor{cvprblue}{rgb}{0.21,0.49,0.74}
\usepackage[pagebackref,breaklinks,colorlinks,citecolor=cvprblue]{hyperref}
\newcommand{\ours}[0]{Creativity-VLA}

\def\paperID{6758} 
\def\confName{CVPR}
\def\confYear{2024}

\title{
Empowering Visual Creativity: \\A Vision-Language Assistant to Image Editing Recommendations
}

\author{
 Tiancheng Shen$^{1,2}$  \hspace{1em}
 Jun Hao Liew$^2$  \hspace{1em}
 Long Mai$^2$  \hspace{1em}
 Lu Qi$^3$  \hspace{1em}
 Jiashi Feng$^2$  \hspace{1em}
 Jiaya Jia$^1$  \\
 $^1$CUHK \hspace{1em} 
 $^2$ByteDance Inc. \hspace{1em}
 $^3$University of California, Merced
}

\begin{document}

\twocolumn[{%
	\maketitle
	\renewcommand\twocolumn[1][]{#1}%
	\begin{center}
		\centering
            \includegraphics[width=1.0\textwidth]{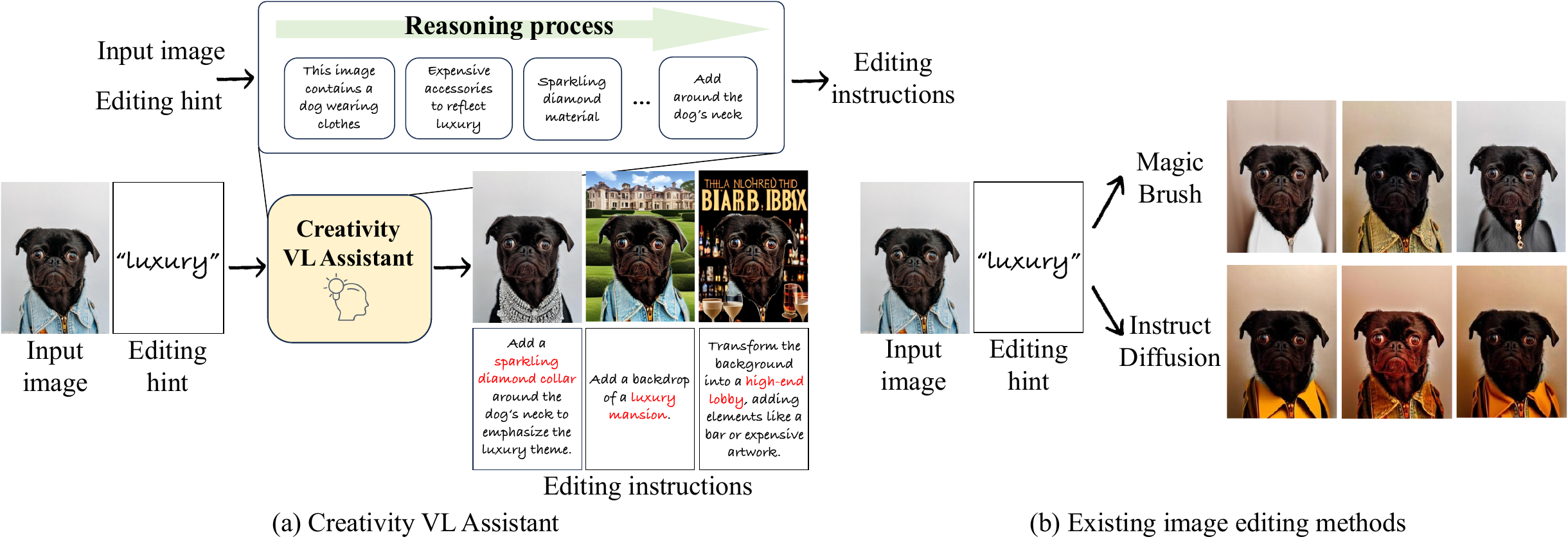}
            \captionof{figure}{            
            In typical editing scenarios, users have the tendency to start with oversimplified text prompts, which we refer to as \textit{editing hint}, as they are often uncertain about what visual results they desire and the corresponding editing instructions needed to achieve appealing editing results. Given only such coarse hints, modern image editing methods~\cite{geng2023instructdiffusion, Zhang2023MagicBrush} often produce unimpressive results, as shown in part~(b). To overcome this, we introduce Creative-VLA which is designed to jointly leverage the visual understanding and creative reasoning capability of Large Vision-Language Models to generate diverse editing instructions, thus achieving the desired visual effect in part (a).
            }
		\label{fig:teaser}
		\vspace{0.2in}
	\end{center}
}]

\begin{abstract}

Advances in text-based image generation and editing have revolutionized content creation, enabling users to create impressive content from imaginative text prompts. However, existing methods are not designed to work well with the oversimplified prompts that are often encountered in typical scenarios when users start their editing with only vague or abstract purposes in mind. Those scenarios demand elaborate ideation efforts from the users to bridge the gap between such vague starting points and the detailed creative ideas needed to depict the desired results. In this paper, we introduce the task of Image Editing Recommendation (IER). This task aims to automatically generate diverse creative editing instructions from an input image and a simple prompt representing the users' under-specified editing purpose. To this end, we introduce Creativity-Vision Language Assistant~(Creativity-VLA), a multimodal framework designed specifically for edit-instruction generation. We train Creativity-VLA on our edit-instruction dataset specifically curated for IER. We further enhance our model with a novel 'token-for-localization' mechanism, enabling it to support both global and local editing operations. Our experimental results demonstrate the effectiveness of \ours{} in suggesting instructions that not only contain engaging creative elements but also maintain high relevance to both the input image and the user's initial hint.

\end{abstract}    
\section{Introduction}
\label{sec:intro}

\begin{figure*}[h]
\begin{center}
\includegraphics[width=0.8\linewidth]{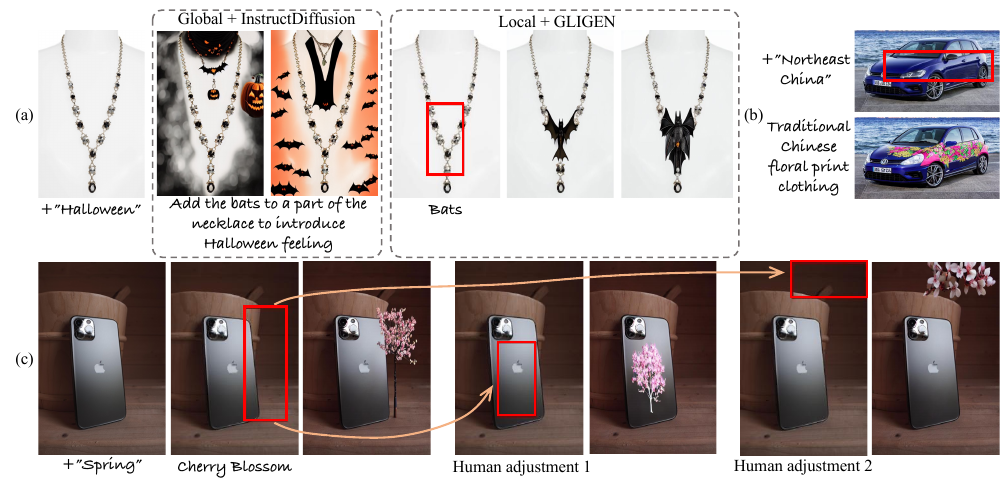}
\end{center}
\vspace{-8mm}
\caption{
The necessity for local editing is highlighted in~(a). InstructDiffusion~\cite{geng2023instructdiffusion} is not well-suited to tasks such as product design that require localized editing models. (b) provides an example of an appropriate region for implementing suggestions. As demonstrated in~(c), decoupling the suggestion and location in the instruction can be beneficial for human adjustments when the predicted location is not ideal and the suggestion is acceptable. The images enclosed by the blue rectangle demonstrate the results of global editing using InstructDiffusion~\cite{geng2023instructdiffusion}, and those within the green rectangle show local editing results achieved using GLIGEN~\cite{Li_2023_CVPR}. The red rectangle indicates the recommended location for the suggestion.
}
\label{fig:local}
\end{figure*} 

Image editing is an integral part of many modern content creation workflows.
From personal hobbyists to professional designers, photo editing enables users to add their touches of imagination and transform ordinary pictures into engaging artworks.
Recent advances in image generation, powered by modern generative models~\cite{rombach2022high, podell2023sdxl, ruiz2023dreambooth}, have rapidly elevated the image editing capabilities beyond low-level enhancement tasks. In particular, many text-based editing systems have been developed to allow users to easily turn their photos into highly imaginative visual content by simply providing textual instructions
~\cite{brooks2023instructpix2pix, geng2023instructdiffusion, Zhang2023MagicBrush, Li_2023_CVPR}.

However, generating creative editing results remains a time-consuming process.
In many practical scenarios, users start their editing with only the source photos along with a vague purpose. 
Generating appealing editing results often requires an elaborate brainstorming and reasoning process to bridge the gap between such a relatively blank slate and detailed, creative, effective editing instructions. 
Such \textit{creativity gap} is especially evident in scenarios where only a high-level theme, mood, or constraint is given.
For instance, when a content creator wants to turn her puppy's photo into one that conveys the theme of a luxury lifestyle, she would start with only the puppy's photo and the specific theme \textit{luxury} in mind (Fig.~\ref{fig:teaser}). In order to get an appealing edited photo, the creator needs to be able to identify and consider various creative ideas (\eg, ``add a sparkling necklace to the dog's neck'') to materialize such high-level purpose. This requires a chain of reasoning process (\eg, analyze the content and composition in the image $\rightarrow$ look for concepts related to the theme $\rightarrow$ choose one of the concepts that fit the image content \etc). Briefly speaking, it requires identifying relevant concepts, reasoning processes, and applying them appropriately to the image content.

Toward narrowing this creativity gap in image editing, we introduce the novel task of Image Editing Recommendation~(IER). IER aims to generate a diverse set of creative editing instructions from a simple prompt representing a user's vague purpose~(which we refer to as \textit{editing hint}) such that the generated instructions align with both the image content and the user's high-level objectives. 
In the example of enhancing a puppy photo with a `luxury' theme, our IER system would suggest plausible creative modifications, such as changing the background luxurious environments or adding a sparkling diamond collar (Fig.~\ref{fig:teaser}(a)).

Addressing the challenges of IER involves developing a system with is capable of: 1) high-level reasoning, 2) visual understanding, and 3) supporting both global and local editing modes. To this end, we present Creativity-Vision Language Assistant (\ours{}), a multimodal framework tailored for IER. 
To jointly address the first two desired capabilities for IER, we leverage a Vision-Language Model (VLM) proficient in generating contextually relevant and visually interesting editing instructions. 
To train our creative-domain specific model, we curate an instruction dataset consisting of 16,000 \{image, editing hint, editing instruction, edited object\} tuples, obtained through a combination of GPT-4 prompted with Chain-of-Thought (CoT)~\cite{wei2022chain, chu2023survey} techniques and extensive manual curation.

To handle the third aforementioned desired capability of IER, \ours{} explicitly decouples the suggested instructions into two output components: text suggestions and locations, which support practical global and local editing scenarios.
More specifically, \ours{} extends beyond global textual editing instruction by introducing the ``token-for-localization" mechanism, which predicts specific image regions for applying editing instructions. This feature facilitates text-driven local editing and enhances the model's versatility. Take designing a ``necklace with Halloween elements'' in Fig.~\ref{fig:local}(a) as an example, \ours{} suggests not only the relevant content (\ie the bats) to add to the image but also where in the image to best apply that addition.



Our work is dedicated to bridging the creativity gap in image editing, offering users without much professional experience the ease of exploring a wider range of creative editing possibilities. We focus on developing an intelligent editing assistant that aids users in not only materializing their creative ideas into final results but also in the initial ideation of these creative concepts. The key contributions of our work are outlined as follows:

\begin{enumerate}
\item \textit{Image Editing Recommendation (IER) Task}. To address the challenge of translating vague editing hints into specific, actionable editing instructions, we introduce the Image Editing Recommendation (IER) task. This task is designed to assist users, particularly those without a clear vision of the desired outcome, by providing them with diverse and practical editing instructions. These suggestions aim to spark creativity and guide users towards satisfying end results.
\item \textit{Creativity Instruction Dataset}. We have developed a unique dataset, compiled through a Chain-of-Thought (CoT) process that emulates human creative reasoning. This dataset serves as the training foundation for a Vision Language Model (VLM) that excels in recommending a variety of creative editing ideas. These ideas are tailored to both the visual content of images and the specific editing hints provided by users.

\item \textit{Creativity-Vision Language Assistant (\ours{})}. We introduce \ours{}, a multimodal framework specifically engineered for the IER task. \ours{} is enhanced with a novel `token-for-localization' feature, enabling it to support both global and local editing operations. Our experimental results validate the efficacy of \ours{} in generating diverse editing instructions. These instructions not only incorporate engaging creative elements but also maintain high relevance to both the input image and the user's initial hint.
\end{enumerate}

\section{Related Work}
\label{sec:relat}

\begin{figure*}[h]\vspace{-0.2in}
\begin{center}
\includegraphics[width=0.9\linewidth]{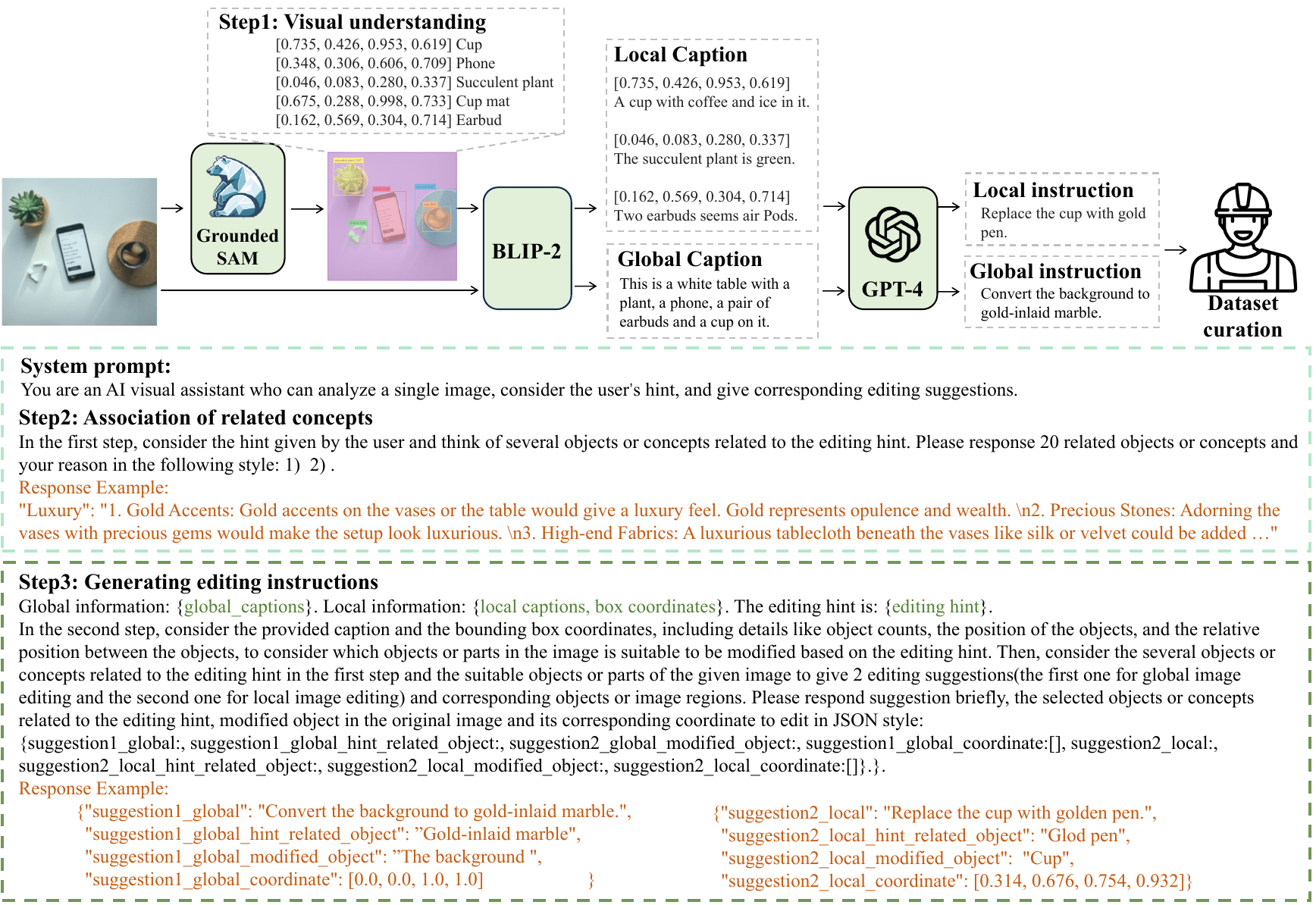}
\end{center}
\vspace{-6mm}
\caption{The pipeline of collecting instruction dataset for imagination in editing. Visual understanding, imagining hint-related concepts, reason to generate instruction and dataset curation form a chain of reasoning processes to obtain high-quality data.}
\label{fig:InstructDataset}
\end{figure*} 

\subsection{Image Editing}
Traditional image editing methods~\cite{gonzales1987digital} often rely on manual operations and professional software such as Adobe Photoshop. Later, certain image editing works to target a single editing function~\cite{huang2018multimodal, gatys2015neural}. Next, several editing algorithms encode the image into latent space and then manipulate latent vectors to edit~\cite{karras2019style}. More recently, researchers applied pre-trained text-to-image~(T2I) diffusion models for image editing~\cite{avrahami2022blended}. For word and image region alignment, Prompt2Prompt~\cite{hertz2022prompt} modifies words in the original prompts to perform editing by cross-attention control. \cite {brooks2023instructpix2pix} improve it to editing instructions. To make the instruction-editing paradigm a unifying and generic framework, InstructDiffusion~\cite{geng2023instructdiffusion} can handle a variety of vision tasks and exhibits the ability to handle unseen user requirements on novel data.

Overall, recent image editing focuses on realizing specific image editing instructions given by users. But in reality, there exists a gap between the initial vague editing hint \vs the detailed instructions used in existing image editing tools. In this paper, we study the IER task to narrow this gap to ease human efforts in imagination.


\subsection{Vision-Language Models} 
Typical VLM, exemplified by CLIP~\cite{radford2021learning} and DALL-E~\cite{ramesh2021zero}, have showcased extraordinary capabilities in understanding and generating cross-modal content. With the advent of the LLM~\cite{touvron2023llama, zhang2022opt}, a growing number of researchers are strategizing the design of VLM. Flamingo~\cite{alayrac2022flamingo} applies gated cross-attention to fuse vision and language modalities. Its special text and vision data format and training strategy enable its in-context learning ability. BLIP-2~\cite{li2023blip} proposes a two-stage strategy to pre-train a Querying Transformer for the alignment between vision and language modalities. MiniGPT-4~\cite{zhu2023minigpt} only uses a linear projection layer to be trained for vision and language modalities alignment. LLaVA~\cite{liu2023llava} follows a similar design principle. For cross-modal tasks, VisionLLM~\cite{wang2023visionllm} fully utilizes instruction tuning to offer a flexible interaction interface for multiple vision-centric tasks. HuBo-VLM~\cite{dong2023hubo} is proposed to tackle perception tasks associated with human-robot interaction. Kosmos-2~\cite{peng2023kosmos} constructs a large-scale dataset to train VLM can ground image-text pairs.

However, these works mainly focus on perceptual tasks~\cite{liu2023llava, liu2023improvedllava, alayrac2022flamingo, wang2023visionllm, dong2023hubo, li2022grounded}.
Differently, we empower \ours{} to generate diverse ideas for tackling the IER problem.

\section{Method}
\label{sec:metho}

Toward narrowing the creativity gap in image editing, we begin with the definition of the IER task. Subsequently, we introduce an instruction dataset for finetuning our proposed \ours{} model. This model can generate a diverse set of editing instructions based on the image content and editing hint. Finally, we delineate the architectural details and functions of \ours{}.


\subsection{Image editing recommendation}

Recent advancements in diffusion models~\cite{song2019generative, rombach2022high} have significantly propelled the evolution of image editing algorithms~\cite{podell2023sdxl, ruiz2023dreambooth, brooks2023instructpix2pix, geng2023instructdiffusion, Zhang2023MagicBrush, Li_2023_CVPR}. These algorithms now enable a broad spectrum of editing applications, including object manipulation, background replacement, object removal, \etc.  Despite the impressive capabilities of these image editing systems, a practical challenge remains: bridging the creativity gap between initial, often vague editing hints and the detailed, creative instructions necessary for these algorithms to function optimally.

In this work, we introduce the Image Editing Recommendation (IER) task. The objective of IER is to generate a diverse set of editing instructions automatically, denoted as $\boldsymbol{O}_{inst}$, derived from a base image $\boldsymbol{I}_{img}$ and a simple editing hint $\boldsymbol{I}_{hint}$. These hints are supposed to encapsulate high-level editing intention. The generated instructions $\boldsymbol{O}_{inst}$ are aimed to be seamlessly integrated into existing image editing systems, facilitating the production of compelling and creative editing outcomes.

Next, we present our instruction dataset for imagination in image editing, which we use to finetune our \ours{} for addressing the IER task.





\subsection{Instruction dataset for imagination in editing}
Since the emergence of LLMs, VLMs have also attracted much research effort~\cite {alayrac2022flamingo, li2023blip, zhu2023minigpt, liu2023llava}. VLMs are often accompanied by the release of cross-modal instruction datasets for target functions~\cite{wang2023visionllm, dong2023hubo, peng2023kosmos}. These datasets focus on image description, question answering, and object relationship reasoning, but few are geared toward creativity. 
To target the IER problem, we collect an instruction dataset to stimulate divergent imagination and generate creative editing suggestions. Inspired by the data generation potential of GPT-4~\cite{openai2023gpt4}, we devise a data collection procedure that utilizes the vision model for dense perception and a dedicated CoT~\cite{wei2022chain, chu2023survey} technique for prompting GPT-4 to collect a multimodal instruction dataset.
The collection process can be decomposed into 4 steps:

\noindent \textbf{Step 1. Visual understanding.}
The first step is to extract an exhaustive list of visual elements present in the image for comprehensive visual understanding. To this end, we employ RAM~\cite{zhang2023recognize}, Grounding-DINO~\cite{liu2023grounding}, and SAM~\cite{kirillov2023segany} to extract open-world category and spatial coordinates of each object in the image. In addition to category information, we also use BLIP-2~\cite{li2023blip} to generate detailed text descriptions for each local object and the entire image. This process helps convert images into textual representations with rich fine-grained details (gray dashed boxes in Fig.~\ref{fig:InstructDataset}).


\noindent \textbf{Step 2. Association of related concepts.}
Given the image content and an abstract hint (\eg, luxury), the second step is to generate a list of hint-related concepts (\eg, a gold necklace, luxury mansion, high-end lobby \etc). To achieve this, we prompt GPT-4~\cite{openai2023gpt4} to imagine around the editing hint. We also prompt it to provide the reasons for each associated concept to ensure it is reasonable. The prompt and examples are in the light green dashed box in Fig.~\ref{fig:InstructDataset}. 


\noindent \textbf{Step 3. Generating editing instructions.}
Given the visual elements and hint-associated concepts in the previous two steps, the third step is to prompt GPT-4~\cite{openai2023gpt4} to produce editing instructions in JSON format. 
Notably, when providing GPT-4~\cite{openai2023gpt4} with in-context examples, we observe a significant reduction in the diversity of the editing actions.
Therefore, we do not use the examples. Instead, we specify the requirements in more detail in the prompt and manually filter and refine the instruction data. The complete prompt used can be found in the supplementary material. We show the simplified prompt in the dark green dashed box in Fig.~\ref{fig:InstructDataset}.


\noindent \textbf{Step 4. Dataset curation.}
We first deduplicate the instruction dataset
with GPT-4~\cite{openai2023gpt4}. Next, because whether the instructions are acceptable for the image editing model~\cite{brooks2023instructpix2pix, geng2023instructdiffusion} is crucial for the IER task, we devise a manual curation process to make sure the instructions that result in failure editing cases are not used in instruction tuning. We perform the manual curation in the following ways: for global editing 
we filter out the instruction, whose editing result's CLIP~\cite{radford2021learning} similarity score (with the instruction itself) is lower than 0.4; for local editing
the instruction should always be simple enough to ensure a clear editing effect, such as ``a corgi". 

In total, there are 16K editing instructions in the dataset, covering human, animal, indoor and outdoor scenes, commodity closeups, \etc Global and local instructions each account for 50\%. Every instruction is converted into the template ``\textit{ HUMAN: $\langle$IMAGE$\rangle$ Given the input image and analyzing the image content, please give one image editing suggestion about the editing hint: \{editing hint\}. ASSISTANT: For global/local editing suggestion: \{specific editing suggestion\}.$\langle$EDIT$\rangle$}" for fine-tuning.

\begin{figure*}[t]
\begin{center}
\includegraphics[width=0.90\linewidth]{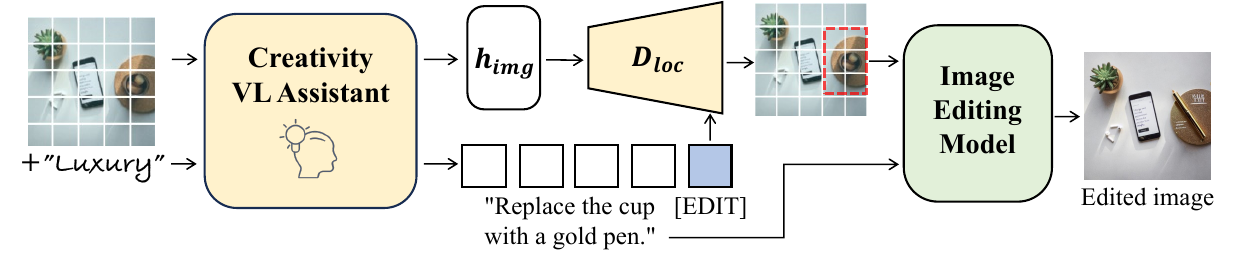}
\end{center}
\vspace{-8mm}
\caption{The architecture of \ours{}. It converts input image and editing hints into editing suggestion and editing token, which is used to recommend the locations for editing.}
\label{fig:MainDesign}
\end{figure*} 

\subsection{Creativity VL Assistant}

To enable \ours{} to generate instructions, we utilize LLaVA~\cite{liu2023llava}, as our backbone VLM denoted as $F$, and finetune it using our instruction dataset. Altogether, its image decoder can perceive image content, and LLM can imagine and reason to generate instructions. 
To support practical image editing, both global and local editing suggestions are required to enable flexible editing.

While global edit instructions are compatible with instruction-following editing models~\cite{brooks2023instructpix2pix, geng2023instructdiffusion, Zhang2023MagicBrush}, these models are often not ideal in expressing local edit instructions. 
A typical example of this is the necklace design showcased in Fig.~\ref{fig:local}, where a bounding box or mask to delineate the editable region results in better than global editing which inevitably introduces too many unwanted modifications for a necklace designer. 
These precise image editing scenarios are common in product design and photo authoring, wherein changes are confined to a specific region without affecting the surrounding areas.



However, common VLM~\cite{alayrac2022flamingo, li2023blip, zhu2023minigpt, liu2023llava} can only output text, which is not ideal for conveying precise location information.
Therefore, we decouple the model output $\boldsymbol{O}_{inst}$ into textual editing action suggestion $\boldsymbol{O}_{sug}$ and editing location $\boldsymbol{O}_{loc}$. Inspired by Query-based visual localization works~\cite{cheng2022masked, qi2023high}, we introduce a special token $\langle EDIT \rangle$
to predict the location recommended for editing by computing cross-attention with the image features.
The editing location $\boldsymbol{O}_{loc}$ works as guidance in image editing models that specialize in local editing, such as GLIGEN~\cite{Li_2023_CVPR} and SDXL-Inpainting~\cite{rombach2022high}. In practice, when the user is not satisfied with the recommended location, the user can easily choose the appropriate location manually to implement the editing suggestions $\boldsymbol{O}_{sug}$. 


\noindent \textbf{Image editing suggestion.}
Following the instruction tuning paradigm in previous VLM works~\cite{liu2023llava,li2023blip}, our \ours{} learns to decompose the problem, think step by step implicitly, and generate diverse editing suggestions $\boldsymbol{O}_{sug}$.

\noindent \textbf{Token for location.}
In order to be able to recommend potential locations for each editing suggestion, we introduce a special token, $\langle EDIT \rangle$ into LLaVA~\cite{liu2023llava}'s vocabulary. This can be formulated as
\begin{equation}
    [\boldsymbol{O}_{sug}, \langle EDIT \rangle]=\boldsymbol{F}\left(\boldsymbol{I}_{img}, \boldsymbol{I}_{hint}\right).
\end{equation}
During the training and generation process, this spatial token appears within the output sequence. $\langle EDIT \rangle$’s corresponding embedding $\boldsymbol{h}_{EDIT}$ in LLM is further processed by a localization decoder $\boldsymbol{D}_{loc}$ consisting of an editing token projection layer and three transformer layers. Similar to a query-based decoder in~\cite{cheng2022masked, qi2023high}, $\boldsymbol{h}_{EDIT}$ calculates with the image feature $\boldsymbol{h}_{img}$ from $\boldsymbol{F}_{enc}$(CLIP~\cite{radford2021learning}'s visual encoder) to generate a bounding box as location recommendation as
\begin{equation}
    \boldsymbol{h}_{img} = \boldsymbol{F}_{enc} \left(\boldsymbol{I}_{img}\right), 
\end{equation}
\begin{equation}
    \boldsymbol{O}_{loc} = \boldsymbol{D}_{loc}(\boldsymbol{h}_{EDIT}, \boldsymbol{h}_{img}). 
\end{equation}


\subsubsection{Training Details}

\noindent \textbf{Trainable parameters.}
We freeze the visual backbone of $F_{enc}$.
The cross-modal projector, the LLM, and location decoders $\boldsymbol{D}_{loc}$ are fine-tuned
to fit the instruction dataset. 

\noindent \textbf{Training objectives.}
The overall model architecture is trained in an end-to-end manner. This process utilizes cross-entropy loss~(CE loss) $\mathcal{L}_{txt}$ for text generation, and localization loss $\mathcal{L}_{loc}$, comprising both L1 loss and generalized intersection over union (GIoU loss) during training. These components are weighted differently \wrt to $\lambda_{txt}$ and $\lambda_{loc}$, respectively.
The composite loss function is defined as follows in the training:
\begin{equation}
\mathcal{L}=\lambda_{txt} \mathcal{L}_{txt}+\lambda_{loc} \mathcal{L}_{loc},
\end{equation}
\vspace{-1mm}
\begin{equation}
\mathcal{L}_{txt} = \mathcal{L}_{CE} \left(\boldsymbol{O}_{sug}, \widehat{\boldsymbol{O}}_{sug} \right), 
\end{equation}
\vspace{-1mm}
\begin{equation}
\mathcal{L}_{loc} = \mathcal{L}_{L1} \left(\widehat{\boldsymbol{O}}_{loc}, \boldsymbol{O}_{loc} \right) +  \mathcal{L}_{GIoU} \left(\widehat{\boldsymbol{O}}_{loc}, \boldsymbol{O}_{loc} \right), 
\end{equation}
where $\boldsymbol{O}_{sug}$ and  $\boldsymbol{O}_{loc}$ are the output of \ours{}, and the corresponding $\widehat{\boldsymbol{O}}_{sug}$ and  $\widehat{\boldsymbol{O}}_{loc}$ are the ground truth.
\section{Experiment}
\label{sec:exper}

\subsection{Experiment Settings}

\textbf{Network architecture.} Our \ours{} incorporates the LLaVA-7B Vision Language Model (VLM) \cite{liu2023llava} as the foundational VLM~($F$). We enhance this with a specialized localization decoder, $\boldsymbol{D}_Oth{loc}$, which consists of an editing token projection layer (configured as an MLP with channels [1024, 1024]) and a 3-layer transformer, in line with methodologies outlined in \cite{cheng2022masked, qi2023high}. The weights of the textual loss ($\lambda_{txt}$) and the localization loss ($\lambda_{loc}$) are set to 1 and 2, respectively. We fine-tune the model using our custom imagination instruction dataset over 3 epochs.  

\noindent \textbf{Image editing tools.} We employ InstructDiffusion~\cite{geng2023instructdiffusion} for global editing and GLIGEN~\cite{Li_2023_CVPR} for local editing. Note that given the generalizability of our method, it also supports other editing tools such as InstructPix2Pix~\cite{brooks2023instructpix2pix}, MagicBrush~\cite{Zhang2023MagicBrush}, SDXL-inpainting~\cite{podell2023sdxl}, \etc




\noindent \textbf{Baselines.} 
We compare our \ours{} with four other methods, categorized into two groups. The first group includes MagicBrush~(MB)~\cite{Zhang2023MagicBrush} and InstructDiffusion~(InsDiff)~\cite{geng2023instructdiffusion}, chosen to assess if the editing instructions from \ours{} can effectively narrow the creativity gap and elevate user satisfaction. The input for both MB and InsDiff is structured as \textit{``Make/Transfer it \{editing hint\}''} or \textit{``Add \{editing hint\} to \{the main object of the image\}''}.

The second group assesses the quality of the recommended editing instructions. We choose LLaVA-v1.5 \cite{liu2023improvedllava} and GPT-4V \cite{gpt4vblog, gpt4v, gpt4vcontribution} as representative VLM competitors. All methods receive identical images and editing hints, with their generated instructions fed into InstructDiffusion \cite{geng2023instructdiffusion} for result generation. For fairness, we include a Chain of Thought (CoT) prompt \cite{wei2022chain, chu2023survey} for LLaVA-v1.5 and GPT-4V, along with a 'one-sentence reply' requirement to conform to InstructDiffusion's input constraints.

\begin{table}[t]
\begin{center}
\small
\setlength{\tabcolsep}{2.5mm}{
\begin{tabular}{lccc}
\hline
\cellcolor{lightgray!30}Method  & \cellcolor{lightgray!30}MB & \cellcolor{lightgray!30}InsDiff & \cellcolor{lightgray!30}\ours{}  \\
\hline
Hint Alignment
                & 2.27 & 2.43 & \textbf{1.31}  \\
Image Alignment
                & \textbf{1.63} & 2.07 & 2.32   \\
Visual quality
                & 2.28 & 2.36 & \textbf{1.33}  \\
Diversity
                & 1.51 & 2.43 & \textbf{1.07} \\
\hline
\end{tabular}}
\vspace{-3mm}
\caption{User preference of MagicBrush, InstructDiffusion, and \ours{} in generation preview with editing hint. 
} 
\vspace{-1em}
\label{table:ver_sug}
\end{center}
\end{table}

\begin{table}[h]
\begin{center}
\small
\setlength{\tabcolsep}{1.5mm}{
\begin{tabular}{lccc}
\hline
\cellcolor{lightgray!30}Method & \cellcolor{lightgray!30} LLaVA-v1.5 & \cellcolor{lightgray!30}GPT-4V & \cellcolor{lightgray!30}\ours{}   \\
\hline
Hint Alignment
               & 2.39 & 2.28  & \textbf{1.33}  \\
Image Alignment
               & 2.34 & 1.94 & \textbf{1.73}   \\
Visual quality
               & 2.52 & 2.11  & \textbf{1.32} \\
Diversity
               & 2.30 & 2.33  & \textbf{1.36} \\
\hline
\end{tabular}}
\vspace{-3mm}
\caption{User preference of LLaVA-v1.5, GPT-4V, and \ours{} cooperating with InstructDiffusion~\cite{geng2023instructdiffusion} for IER. 
} 
\vspace{-2em}
\label{table:vlm_comp}
\end{center}
\end{table}

\subsubsection{User Preference Evaluation}
\label{sec:userstudy}
We begin with a user study. 
The participants are requested to rank the outputs of these models (the lower the better) based on the following four metrics:
(1) Hint Alignment: the consistency between the editing hint and the results;
(2) Image Alignment: the consistency between the input image and the editing result;
(3) Visual Quality: the aesthetic appeal, satisfaction, and inspirational value of the editing results;
(4) Diversity: diversity within the editing results.

\begin{figure*}[t]\vspace{-0.0in}
\begin{center}
\includegraphics[width=0.8\linewidth]{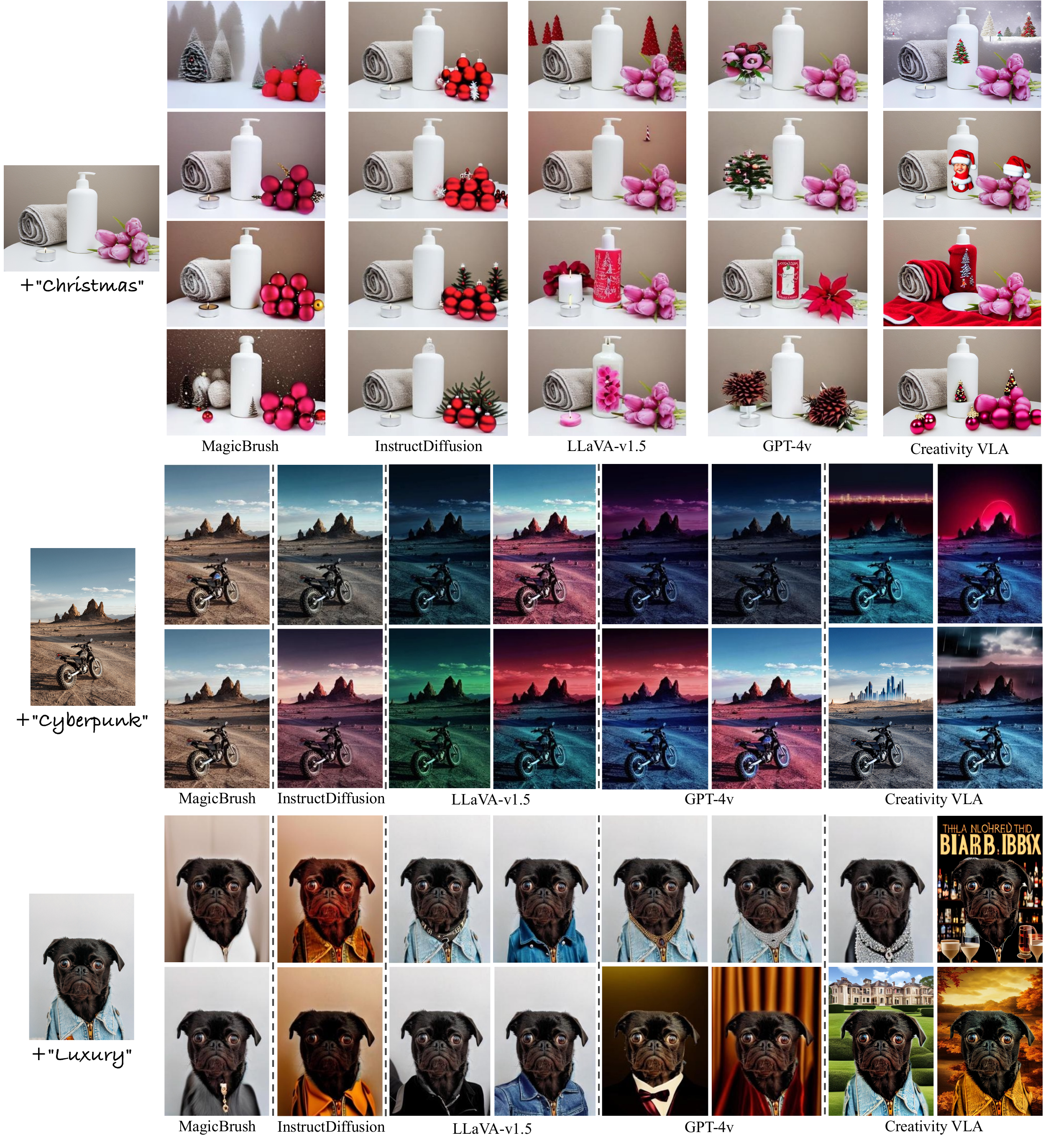}
\end{center}
\vspace{-6mm}
\caption{Qualitative comparison among MagicBrush, InstructDiffusion, LLaVA-v1.5, GPT-4V and \ours{}. Due to the space limitation, corresponding editing instructions are in the supplementary file.}
\label{fig:Qualitative_comparisons}
\end{figure*}

Results in Table~\ref{table:ver_sug} indicate that \ours{} consistently outperforms in editing hint alignment, visual quality, and diversity. This suggests our model works better in effectively bridging the creativity gap and eases the burden of imagination on users. However, image alignment scores inversely correlate with other metrics, likely due to substantial modifications being conducted.
On the other hand, as shown in Fig.~\ref{fig:Qualitative_comparisons}, other methods do not make large modifications, which leads to higher image alignment scores.

Table~\ref{table:vlm_comp} shows that \ours{} outperforms both LLaVA-v1.5 and GPT-4V, despite our instruction dataset is generated from GPT-4 \cite{radford2019language}. This is attributed to the inclusion of CoT manner in data collection and meticulous dataset curation, enhancing instruction's effectiveness. Notably, despite guidance on CoT and output length, the outputs of  LLaVA-v1.5 and GPT-4V do not convert well into editing results via InstructDiffusion~\cite{geng2023instructdiffusion} and have duplication (see Fig.~\ref{fig:Qualitative_comparisons}).

\begin{table}[t]
\begin{center}
\small 
\setlength{\tabcolsep}{1.2mm}{
\begin{tabular}{lccccc}
\hline
\cellcolor{lightgray!30}Alignment  & \cellcolor{lightgray!30}MB & \cellcolor{lightgray!30}InsDiff & \cellcolor{lightgray!30} LLaVA-v1.5 & \cellcolor{lightgray!30}GPT-4V  & \cellcolor{lightgray!30}C-VLA  \\
\hline
Hint 
                & 0.142 &  0.143 & 0.144  & 0.154 & \textbf{0.172} \\
Image 
                & \textbf{0.918} & 0.922 & 0.860  & 0.854 & 0.842 \\
\hline
\end{tabular}}
\vspace{-3mm}
\caption{CLIP similarity scores \wrt the input image and editing hint. C-VLA represents \ours{} here.} 
\vspace{-1em}
\label{table:clipsim}
\end{center}
\end{table}

\subsubsection{CLIP score evaluation}

The CLIP score evaluation measures the similarity between the editing results from all five methods \wrt both the input image and editing hint (Table~\ref{table:clipsim}). \ours{} demonstrates competitive performance in image alignment, while outperforming all others in hint alignment.

\begin{figure*}[!t]\vspace{-0.1in}
\begin{center}
\includegraphics[width=0.99\linewidth]{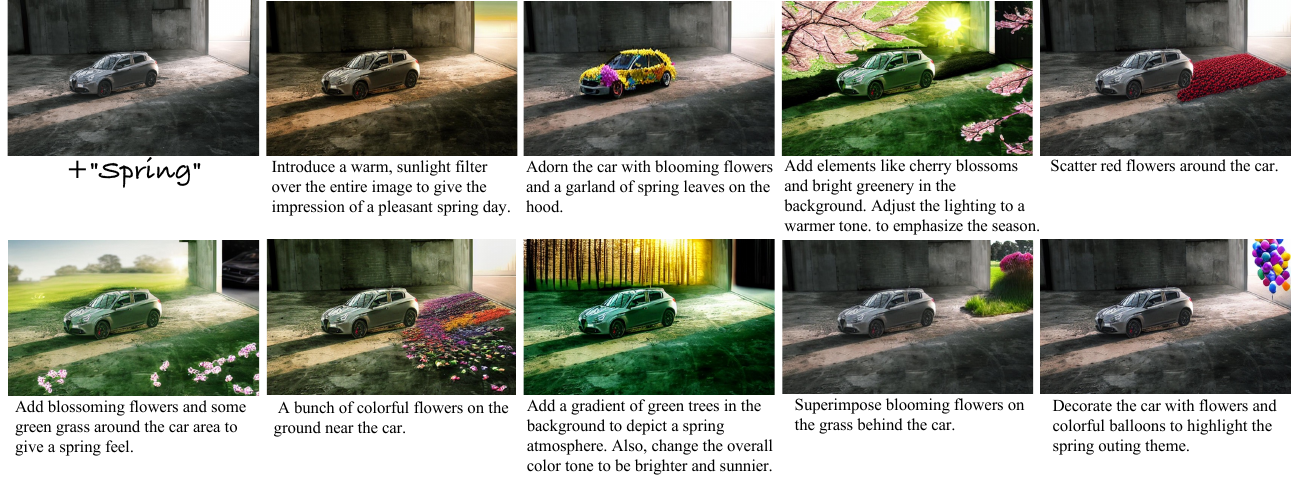}
\end{center}
\vspace{-8mm}
\caption{Example to show the diversity of \ours{}. 
This example shows that our \ours{} is beneficial for inspiring creativity (\eg, when designing the spring season's advertisements to attract customers' attention)
}
\label{fig:diversity}
\end{figure*}

\begin{figure*}[h]
\begin{center}
\includegraphics[width=0.99\linewidth]{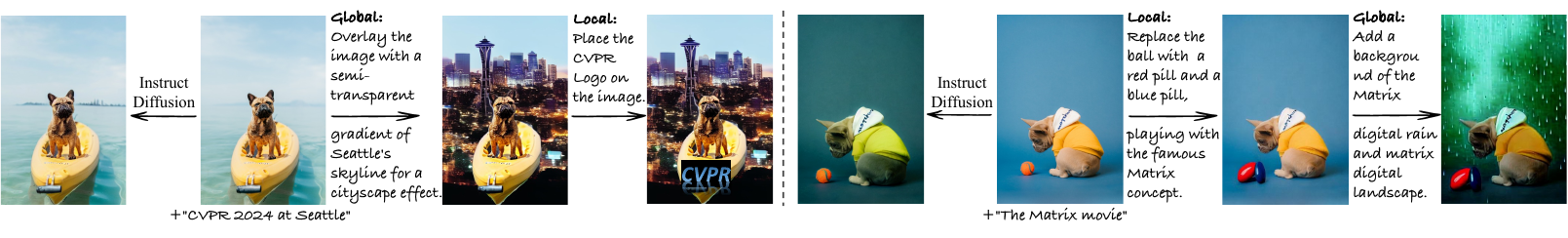}
\end{center}
\vspace{-8mm}
\caption{Examples of sequential editing with \ours{}. We show that local and global editing can work collaboratively. CVPR logo is not generated with GLIGEN~\cite{Li_2023_CVPR}.}
\label{fig:sequence}
\end{figure*}

\subsubsection{Qualitative Comparisons}

Qualitative comparisons are shown in Fig.~\ref{fig:Qualitative_comparisons}, with additional results provided in the supplementary materials. Those examples demonstrate that \ours{} adeptly incorporates relevant and meaningful concepts related to the provided hints into the editing instructions. This leads to more practical and contextually appropriate outcomes. In contrast, other methods show limited diversity, often introducing irrelevant objects or merely adjusting global colors. With \ours{}, users can edit images with vague hints while obtaining diverse plausible editing recommendations and sparkling results.

\subsubsection{How may \ours{} assist humans?}
We use the results above to highlight some cases where users might find \ours{} helpful in image design and editing. Based on the feedback from user study, in general, this model can assist users in the following ways:
\begin{enumerate}
\item Enhance creativity. Recommends diverse design options and perspectives, expanding creative possibilities and sparking new inspirations.
\item  Quickly transforms initial ideas into previews, shortening the time from coarse concept to prototype.
\item Iterative improvement. Decoupling suggestions and locations allows for fine-tuning based on model-generated previews, enhancing the quality of the final work.
\item Lower technical barriers. Enables non-professional users to get diverse instructions and visual effects, liberating people who are struggling with creativity.
\end{enumerate}

\subsection{Ablation study}
Here we do ablation studies to verify the diversity of \ours{} in Fig.~\ref{fig:diversity}, and use instructions in sequence to realize a more appealing editing effect in Fig.~\ref{fig:sequence}. These examples all prove that our method is valuable in generating diverse instructions efficiently and performing more in-depth editing.
More innovative and intriguing results are detailed in the supplementary materials.




\section{Conclusion}
\label{sec:concl}



In conclusion, our work contributes to narrowing the creativity gap in image editing, empowering users without expansive professional experience to explore and actualize their creative visions. By introducing the Image Editing Recommendation (IER) task, we establish a framework that effectively translates vague editing hints into precise, actionable instructions. This not only assists users in realizing their desired outcomes but also catalyzes the creative process, particularly for those who may start with only a coarse concept. Our development of the Creativity Instruction Dataset, grounded in Chain-of-Thought processes, has been instrumental in training our VLM to produce diverse and contextually relevant editing ideas. Most notably, our \ours{} stands as a testament to our commitment to enhancing the image editing domain. 
Its unique 'token-for-localization' feature empowers users with the ability to perform both global and local edits, thereby increasing the versatility and applicability of our system. 
The experimental results demonstrate the effectiveness of \ours{} in generating varied editing instructions that are not only creatively stimulating but also highly pertinent to both the input image and the user's initial hints. Through this work, we aspire to contribute a new possibility to the landscape of image editing, making it a more accessible and creatively fulfilling endeavor for a broad spectrum of users.





{
    \small
    \bibliographystyle{ieeenat_fullname}
    \bibliography{main}
}


\end{document}